\documentclass[10pt,twocolumn,letterpaper]{article}

\usepackage{cvpr}
\usepackage{times}
\usepackage{epsfig}
\usepackage{graphicx}
\usepackage{amsmath}
\usepackage{amssymb}
\usepackage{subfigure}
\usepackage{color}
\usepackage{multirow}
\usepackage{cite}
\usepackage{epstopdf}
\usepackage{pifont}
\usepackage[dvipsnames]{xcolor}
\usepackage{caption}
\graphicspath{{figures/}}

\def\figvspace{{\vspace{-4mm}}}
\newcommand{\Paragraph}[1]{\vspace{-0mm} \noindent \textbf{#1} \hspace{0mm}}

\newcommand*{\boxedcolor}{red}
\makeatletter
\renewcommand{\boxed}[1]{\textcolor{\boxedcolor}{%
  \fbox{\normalcolor\m@th$\displaystyle#1$}}}
\makeatother

\makeatletter
  \newcommand\figcaption{\def\@captype{figure}\caption}
  \newcommand\tabcaption{\def\@captype{table}\caption}
\makeatother

\usepackage[pagebackref=true,breaklinks=true,letterpaper=true,colorlinks,bookmarks=false]{hyperref}

\cvprfinalcopy % *** Uncomment this line for the final submission

\def\cvprPaperID{****} % *** Enter the CVPR Paper ID here

\ifcvprfinal\fi
\begin{document}

%%%%%%%%% TITLE
\title{DSFD: Dual Shot Face Detector}

\author{Jian Li$^{\dag}$ ~ ~ Yabiao Wang$^{\ddag}$ ~ ~ Changan Wang$^\ddag$ ~ ~ Ying Tai$^\ddag$ \\
~ ~ Jianjun Qian${^{\dag *}}$ ~ ~ Jian Yang${^{\dag *}}$ ~ ~ Chengjie Wang$^\ddag$ ~ ~ Jilin Li$^\ddag$ ~ ~ Feiyue Huang$^\ddag$ \\
\normalsize $^\dag$PCA Lab, Key Lab of Intelligent Perception and Systems for High-Dimensional Information of Ministry of Education\\
\normalsize $^\dag$Jiangsu Key Lab of Image and Video Understanding for Social Security \\
\normalsize $^\dag$School of Computer Science and Engineering, Nanjing University of Science and Technology, Nanjing, China\\
\normalsize $^\ddag$Youtu Lab, Tencent \\
{\tt\small $^\dag$lijiannuist@gmail.com, \{csjqian, csjyang\}@njust.edu.cn} \\
{\tt\small $^\ddag$\{casewang, changanwang, yingtai, jasoncjwang, jerolinli, garyhuang\}@tencent.com}
%{\small \url{https://github.com/TencentYoutuResearch/FaceDetection-DSFD}}
}

%% omit the following lines up until the closing ``}''.
%% Additional authors and addresses can be added with ``\and'',
%% just like the second author.
%% To save space, use either the email address or home page, not both
%\and

%\maketitle

\twocolumn[{%
\renewcommand\twocolumn[1][]{#1}%
\maketitle
\begin{center}
    \centering
\vspace{-5mm}
    \includegraphics[trim={0 0 0 0mm},clip,width=0.9\textwidth]{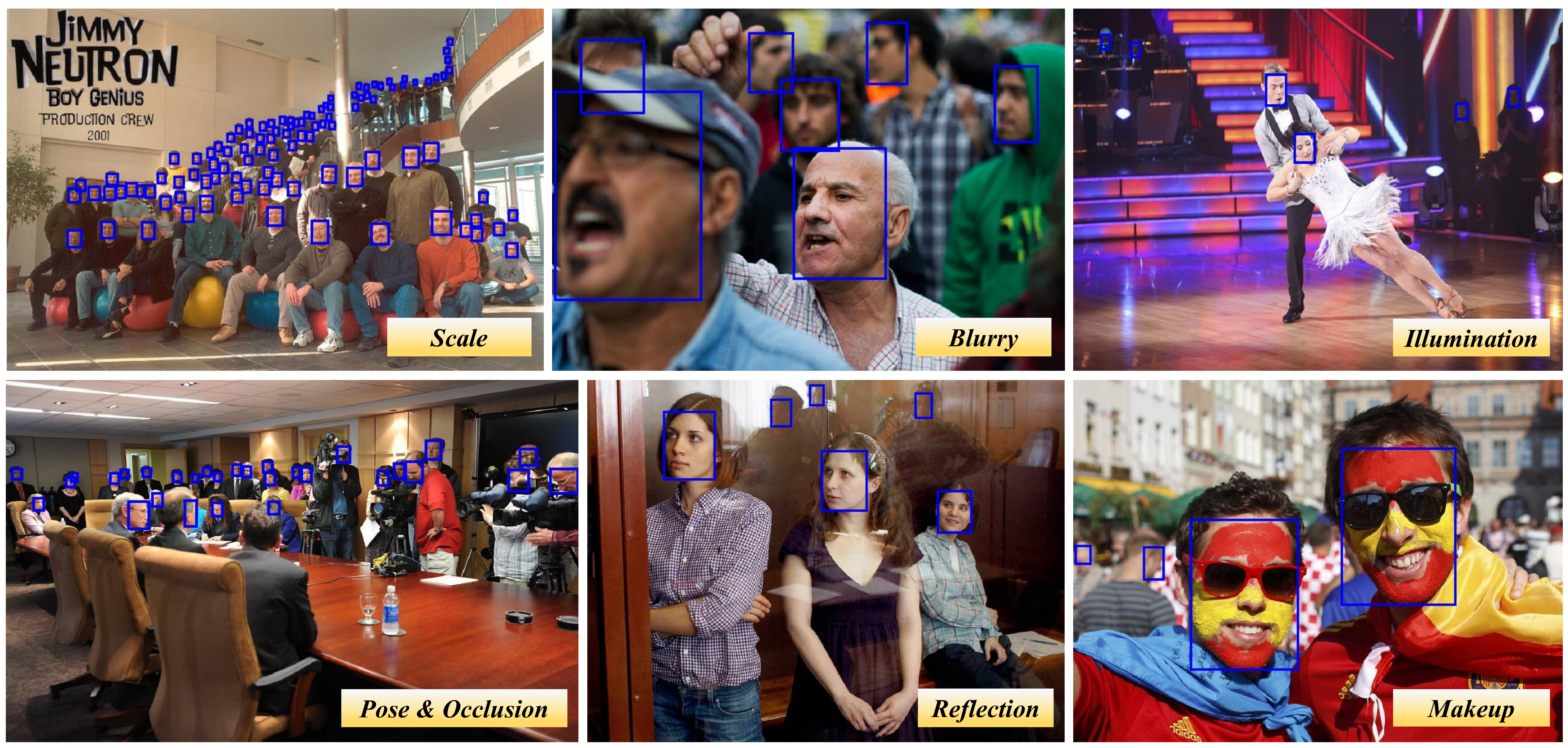} \vspace{-2mm}
    \captionof{figure}{\small \textbf{Visual results}.
    Our method is robust to various variations on scale, blurry, illumination, pose, occlusion, reflection and makeup.}
    \label{fig:selfish} %\figvspace
\end{center}%
}]

{
\renewcommand{\thefootnote}{\fnsymbol{footnote}}
\footnotetext[1]{Jianjun Qian and Jian Yang are corresponding authors.
This work was supported by the National Science Fund of China under Grant Nos. 61876083, U1713208, and Program for Changjiang Scholars.
This work was done when Jian Li was an intern at Tencent Youtu Lab.
}
}

%\maketitle
%\thispagestyle{empty}

%%%%%%%%% ABSTRACT
\begin{abstract}
   %Recently, Convolutional Neural Network (CNN) has achieved great success in face detection.
   %However, it remains a challenging problem for the current face detection methods owing to high degree of variability in scale, pose, occlusion, expression, appearance and illumination.
   In this paper, we propose a novel face detection network with three novel contributions that address three key aspects of face detection, including better feature learning, progressive loss design and anchor assign based data augmentation, respectively.
   First, we propose a Feature Enhance Module (FEM) for enhancing the original feature maps to extend the single shot detector to dual shot detector.
   Second, we adopt Progressive Anchor Loss (PAL) computed by two different sets of anchors to effectively facilitate the features.
   Third, we use an Improved Anchor Matching (IAM) by integrating novel anchor assign strategy into data augmentation to provide better initialization for the regressor.
   Since these techniques are all related to the two-stream design, we name the proposed network as Dual Shot Face Detector (DSFD).
   Extensive experiments on popular benchmarks, WIDER FACE and FDDB, demonstrate the superiority of DSFD over the state-of-the-art face detectors. %(\emph{e.g.}, PyramidBox and SRN).
\end{abstract}

\vspace{-1mm}
%%%%%%%%% BODY TEXT
\section{Introduction}\label{section:1}
Face detection is a fundamental step for various facial applications, like face alignment~\cite{tai-FHR-2019}, parsing~\cite{CT-FSRNet-2018}, recognition~\cite{NMR_PAMI16}, and verification~\cite{deng2018arcface}.
As the pioneering work for face detection, Viola-Jones~\cite{viola2004robust} adopts AdaBoost algorithm with hand-crafted features, which are now replaced by deeply learned features from the convolutional neural network (CNN)~\cite{he2016deep} that achieves great progress.
Although the CNN based face detectors have being extensively studied, detecting faces with high degree of variability in scale, pose, occlusion, expression, appearance and illumination in real-world scenarios remains a challenge.

%\begin{figure}[t]
%  % Requires \usepackage{graphicx}
%  \centering
%  \includegraphics[trim={0 0 0 0mm},clip,width=1\linewidth]{./figures/selfish.pdf}
%  \vspace{-6mm}
%  \caption{\small Our DSFD can find $\textbf{861}$ faces out of the $1000$ facial images present in the above image.
%  } %FIXME the last sentence? (Yes)
%  \label{fig:selfish} \figvspace
%\end{figure}

\begin{figure*}[htp]
  \centering
  \includegraphics[trim={0 0 0 0mm},clip,width=0.95\linewidth]{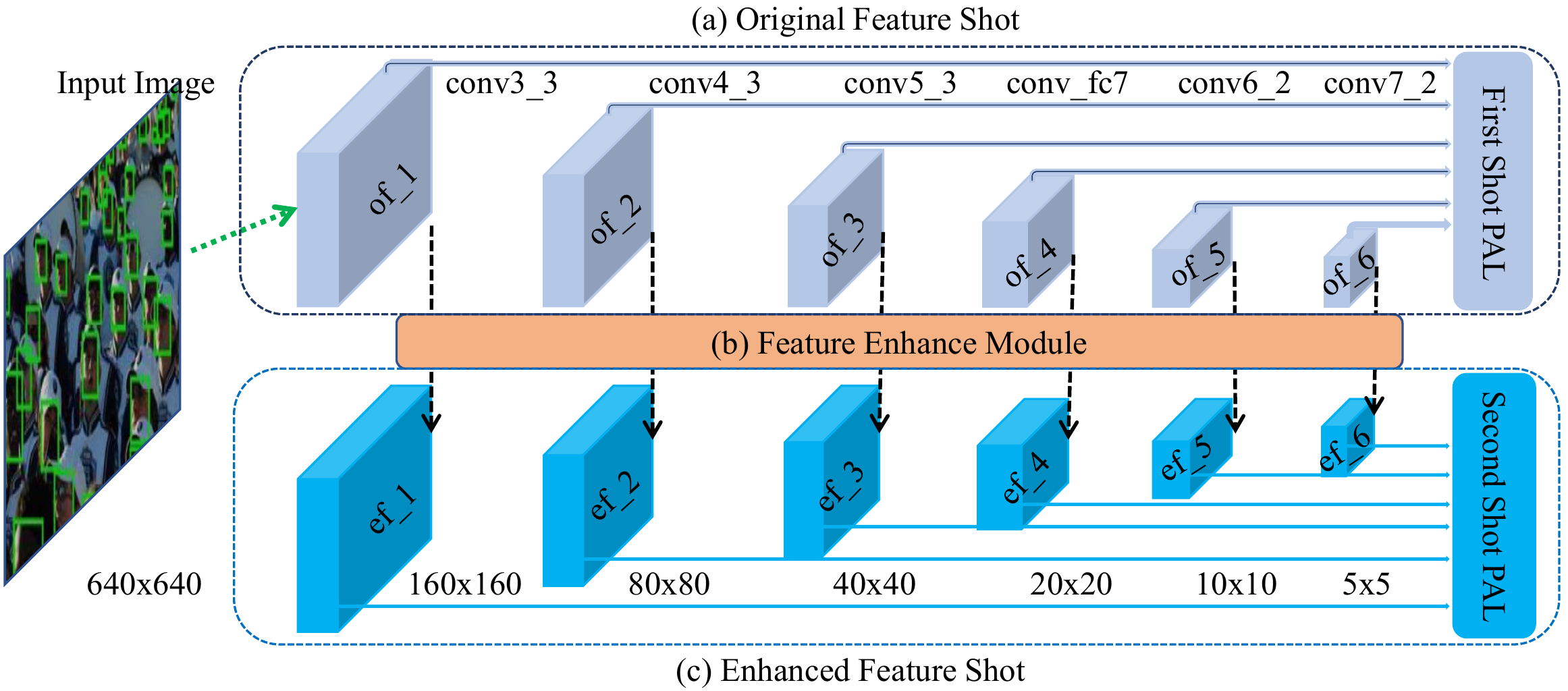}
  \vspace{-3mm}
  \caption{\small \textbf{Our DSFD framework} uses a Feature Enhance Module (b) on top of a feedforward VGG/ResNet architecture to generate the enhanced features (c) from the original features (a), along with two loss layers named first shot PAL for the original features and second shot PAL for the enchanted features.
  } %FIXME the last sentence? (Yes)
  \label{fig:network} \figvspace
\end{figure*}

Previous state-of-the-art face detectors can be roughly divided into two categories.
The first one is mainly based on the Region Proposal Network (RPN) adopted in Faster RCNN~\cite{ren2015faster} and employs two stage detection schemes~\cite{zhang2018face,wang2017facer,wang2017detecting}.
RPN is trained end-to-end and generates high-quality region proposals which are further refined by Fast R-CNN detector.
The other one is Single Shot Detector (SSD)~\cite{liu2016ssd} based one-stage methods, which get rid of RPN, and directly predict the bounding boxes and confidence~\cite{zhang2017s,tang2018pyramidbox,chi2018selective}.
Recently, one-stage face detection framework has attracted more attention due to its higher inference efficiency and straightforward system deployment.

Despite the progress achieved by the above methods, there are still some problems existed in three aspects:

\Paragraph{Feature learning}
Feature extraction part is essential for a face detector.
Currently, Feature Pyramid Network (FPN)~\cite{lin2017feature} is widely used in state-of-the-art face detectors for rich features.
However, FPN just aggregates hierarchical feature maps between high and low-level output layers, which does not consider the current layer's information, and the context relationship between anchors is ignored.

\Paragraph{Loss design}
The conventional loss functions used in object detection include a regression loss for the face region and a classification loss for identifying if a face is detected or not.
To further address the class imbalance problem, Lin \textit{et al.}~\cite{lin2017focal} propose Focal Loss to focus training on a sparse set of hard examples.
To use all original and enhanced features, Zhang \textit{et al.} propose Hierarchical Loss to effectively learn the network~\cite{zhang2017feature}.
However, the above loss functions do not consider progressive learning ability of feature maps in both of different levels and shots.

\Paragraph{Anchor matching}
Basically, pre-set anchors for each feature map are generated by regularly tiling a collection of boxes with different scales and aspect ratios on the image.
Some works~\cite{zhang2017s,tang2018pyramidbox} analyze a series of reasonable anchor scales and anchor compensation strategy to increase positive anchors.
However, such strategy ignores random sampling in data augmentation, which still causes imbalance between positive and negative anchors.
%Continuous face scale and a large number of discrete anchor scales still make huge ratio differences of negative and positive anchors.

In this paper, we propose three novel techniques to address the above three issues, respectively.
%To address the above three issues, we propose  a novel network named Dual Shot Face Detection (DSFD).
First, we introduce a Feature Enhance Module (FEM) to enhance the discriminability and robustness of the features, which combines the advantages of the FPN in PyramidBox and Receptive Field Block (RFB) in RFBNet~\cite{liu2017receptive}.
Second, motivated by the hierarchical loss~\cite{zhang2017feature} and pyramid anchor~\cite{tang2018pyramidbox} in PyramidBox,
%we propose Progressive Anchor Loss (PAL) that computes auxiliary supervision loss by a set of smaller anchors to effectively facilitate the orginal features, since smaller anchor tiled to original feature maps cell may have more semantic information for classification and high-resolution location information for small faces.
we design Progressive Anchor Loss (PAL) that uses progressive anchor sizes for not only different levels, but also different shots.
Specifically, we assign smaller anchor sizes in the first shot, and use larger sizes in the second shot.
Third, we propose Improved Anchor Matching (IAM), which integrates anchor partition strategy and anchor-based data augmentation to better match anchors and ground truth faces, and thus provides better initialization for the regressor.
The three aspects are \textit{complementary} so that these techniques can work together to further improve the performance.
Besides, since these techniques are all related to two-stream design, we name the proposed network as Dual Shot Face Detector (DSFD).
Fig.~\ref{fig:selfish} shows the effectiveness of DSFD on various variations, especially on extreme small faces or heavily occluded faces.

In summary, the main contributions of this paper include:

$\bullet$ A novel Feature Enhance Module to utilize different level information and thus obtain more discriminability and robustness features.

$\bullet$ Auxiliary supervisions introduced in early layers via a set of smaller anchors to effectively facilitate the features.

$\bullet$ An improved anchor matching strategy to match anchors and ground truth faces as far as possible to provide better initialization for the regressor.

$\bullet$ Comprehensive experiments conducted on popular benchmarks FDDB and WIDER FACE to demonstrate the superiority of our proposed DSFD network compared with the state-of-the-art methods.

%-------------------------------------------------------------------------
\section{Related work}\label{section:2}

We review the prior works from three perspectives.

\Paragraph{Feature Learning}
Early works on face detection mainly rely on hand-crafted features, such as Harr-like features~\cite{viola2004robust}, control point set~\cite{abramson2007yet}, edge orientation histograms~\cite{levi2004learning}.
However, hand-crafted features design is lack of guidance. With the great progress of deep learning, hand-crafted features have been replaced by Convolutional Neural Networks (CNN).
For example, Overfeat~\cite{sermanet2013overfeat}, Cascade-CNN~\cite{li2015convolutional}, MTCNN~\cite{zhang2016joint} adopt CNN as a sliding window detector on image pyramid to build feature pyramid.
However, using an image pyramid is slow and memory inefficient. As the result, most two stage detectors extract features on single scale.
R-CNN~\cite{girshick2014rich,girshick2015fast} obtains region proposals by selective search~\cite{uijlings2013selective}, and then forwards each normalized image region through a CNN to classify.
Faster R-CNN~\cite{ren2015faster}, R-FCN~\cite{dai2016r} employ Region Proposal Network (RPN) to generate initial region proposals.
Besides, ROI-pooling~\cite{ren2015faster} and position-sensitive RoI pooling~\cite{dai2016r} are applied to extract features from each region.

More recently, some research indicates that multi-scale features perform better for tiny objects. Specifically, SSD~\cite{liu2016ssd}, MS-CNN~\cite{cai2016unified}, SSH~\cite{najibi2017ssh}, S3FD~\cite{zhang2017s} predict boxes on multiple layers of feature hierarchy.
FCN~\cite{long2015fully}, Hypercolumns~\cite{hariharan2015hypercolumns}, Parsenet~\cite{liu15parsenet} fuse multiple layer features in segmentation.
FPN~\cite{lin2017feature,li2017object}, a top-down architecture, integrate high-level semantic information to all scales.
FPN-based methods, such as FAN~\cite{wang2017face}, PyramidBox~\cite{tang2018pyramidbox} achieve significant improvement on detection.
However, these methods do not consider the current layer¡¯s information.
Different from the above methods that ignore the context relationship between anchors, we propose a feature enhance module that incorporates multi-level dilated convolutional layers to enhance the semantic of the features.

\Paragraph{Loss Design}
Generally, the objective loss in detection is a weighted sum of classification loss (\emph{e.g.} softmax loss) and box regression loss (\emph{e.g.} $L_2$ loss).
Girshick \textit{et al.}~\cite{girshick2015fast} propose smooth $L_1$ loss to prevent exploding gradients.
Lin \textit{et al.}~\cite{lin2017focal} discover that the class imbalance is one obstacle for better performance in one stage detector, hence they propose focal loss, a dynamically scaled cross entropy loss.
Besides, Wang \textit{et al.}~\cite{wang2017repulsion} design RepLoss for pedestrian detection, which improves performance in occlusion scenarios.
FANet~\cite{zhang2017feature} create a hierarchical feature pyramid and presents hierarchical loss for their architecture.
However, the anchors used in FANet are kept the same size in different stages. In this work, we adaptively choose different anchor sizes in different stages to facilitate the features.

\Paragraph{Anchor Matching}
To make the model more robust, most detection methods~\cite{liu2016ssd,zhang2017s,yang2016wider} do data augmentation, such as color distortion, horizontal flipping, random crop and multi-scale training.
Zhang \textit{et al.}~\cite{zhang2017s} propose an anchor compensation strategy to make tiny faces to match enough anchors during training.
Wang \textit{et al.}~\cite{yang2016wider} propose random crop to generate large number of occluded faces for training.
However, these methods ignore random sampling in data augmentation, while ours combines anchor assign to provide better data initialization for anchor matching.

\begin{figure}[t]
  \centering
  \includegraphics[trim={0 0 0 0mm},clip,width=0.95\linewidth]{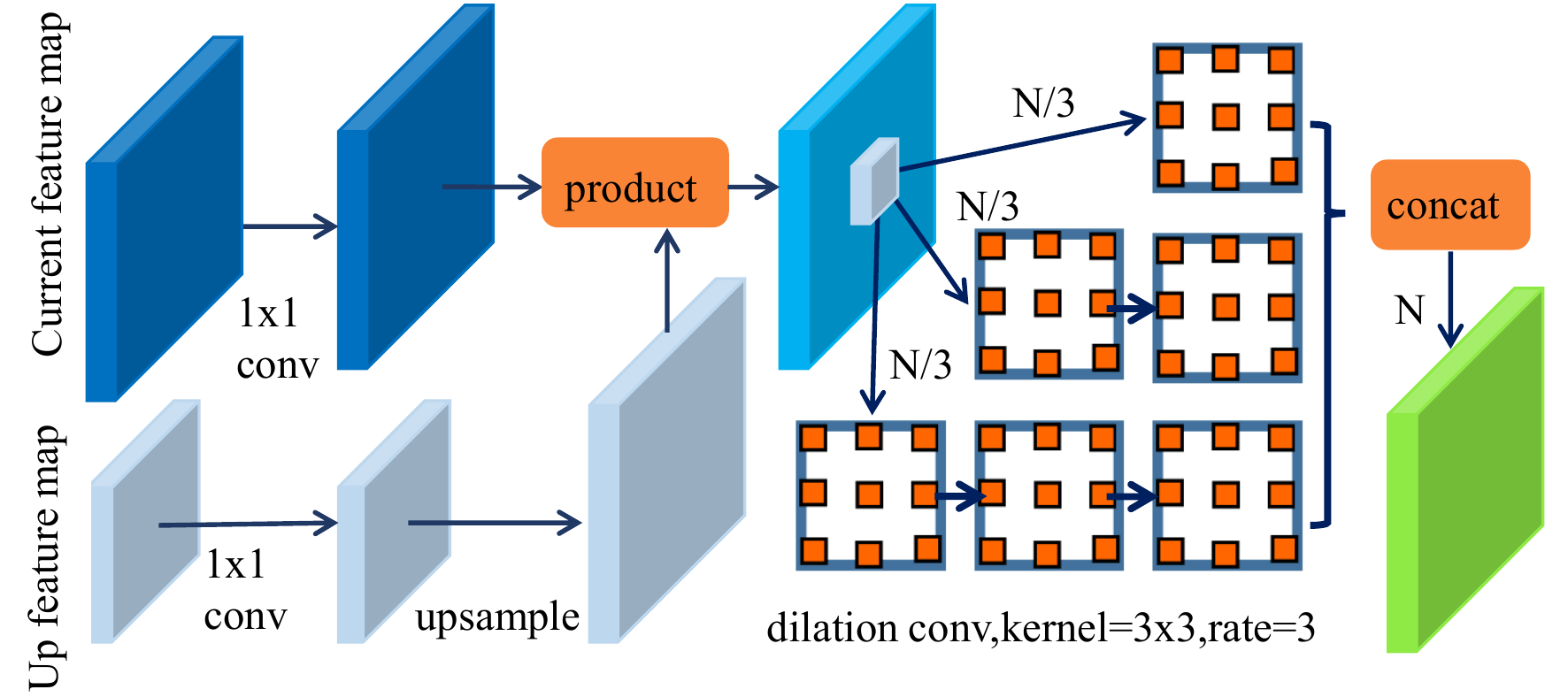}
  \vspace{-1mm}
  \caption{\small \textbf{Illustration on Feature Enhance Module}, in which the current feature map cell interactives with neighbors in current feature maps and up feature maps.
  }
  \label{fig:fem} \figvspace
\end{figure}

\section{Dual Shot Face Detector}\label{section:3}

We firstly introduce the pipeline of our proposed framework DSFD, and then detailly describe our feature enhance module in Sec.~\ref{section:3.2}, progressive anchor loss in Sec.~\ref{section:3.3} and improved anchor matching in Sec.~\ref{section:3.4}, respectively.

\subsection{Pipeline of DSFD}\label{section:3.1}
The framework of DSFD is illustrated in Fig.~\ref{fig:network}.
Our architecture uses the same extended VGG16 backbone as PyramidBox~\cite{tang2018pyramidbox} and S3FD~\cite{zhang2017s}, which is truncated before the classification layers and added with some auxiliary structures.
We select conv$3\_3$, conv$4\_3$, conv$5\_3$, conv$\_$fc$7$, conv$6\_2$ and conv$7\_2$ as the first shot detection layers to generate six original feature maps named $of_1, of_2, of_3, of_4, of_5, of_6$. Then, our proposed FEM transfers these original feature maps into six enhanced feature maps named $ef_1, ef_2, ef_3, ef_4, ef_5, ef_6$, which have the same sizes as the original ones and are fed into SSD-style head to construct the second shot detection layers.
Note that the input size of the training image is $640$, which means the feature map size of the lowest-level layer to highest-level layer is from $160$ to $5$.
Different from S$3$FD and PyramidBox, after we utilize the receptive field enlargement in FEM and the new anchor design strategy, it¡¯s unnecessary for the three sizes of stride, anchor and receptive field to satisfy equal-proportion interval principle.
Therefore, our DSFD is more flexible and robustness. Besides, the original and enhanced shots have two different losses, respectively named First Shot progressive anchor Loss (FSL) and Second Shot progressive anchor Loss (SSL).

%-------------------------------------------------------------------------
\subsection{Feature Enhance Module}\label{section:3.2}
Feature Enhance Module is able to enhance original features to make them more discriminable and robust, which is called FEM for short.
For enhancing original neuron cell $oc_{(i,j,l)}$, FEM utilizes different dimension information including upper layer original neuron cell $oc_{(i,j,l)}$ and current layer non-local neuron cells: $nc_{(i-\varepsilon,j-\varepsilon,l)}$, $nc_{(i-\varepsilon,j,l)}$, ..., $nc_{(i,j+\varepsilon,l)}$, $nc_{(i+\varepsilon,j+\varepsilon,l)}$.
Specially, the enhanced neuron cell $ec_{(i,j,l)}$ can be mathematically defined as follow:
\begin{equation}\
%\small
\begin{aligned}
  ec_{(i,j,l)} &= f_{concat}(f_{dilation}(nc_{(i,j,l)})) \\
  nc_{i,j,l} &= f_{prod}(oc_{(i,j,l)},f_{up}(oc_{(i,j,l+1)}))
  \end{aligned}
  \label{eq:e1}
\end{equation}
where $c_{i,j,l}$ is a cell located in $(i,j)$ coordinate of the feature maps in the $l$-th layer, $f$ denotes a set of basic dilation convolution, elem-wise production, up-sampling or concatenation operations.
Fig.~\ref{fig:fem} illustrates the idea of FEM, which is inspired by FPN~\cite{lin2017feature} and RFB~\cite{liu2017receptive}.
Here, we first use $1$$\times$$1$ convolutional kernel to normalize the feature maps.
Then, we up-sample upper feature maps to do element-wise product with the current ones.
Finally, we split the feature maps to three parts, followed by three sub-networks containing different numbers of dilation convolutional layers.

\subsection{Progressive Anchor Loss}\label{section:3.3}
Different from the traditional detection loss, we design \textit{progressive} anchor sizes for not only different levels, but also different shots in our framework.
Motivated by the statement in~\cite{ren2015faster} that low-level features are more suitable for small faces, we assign smaller anchor sizes in the first shot, and use larger sizes in the second shot.
%In this subsection, we adopt the multi-task loss~\cite{ren2015faster,liu2016ssd} since it helps to facilitate the original and enhanced feature maps training task in two shots.
First, our Second Shot anchor-based multi-task Loss function is defined as:
\begin{equation}\
%\small
\begin{aligned}
  \mathcal{L}_{SSL}(p_i,p^*_i,t_i,g_i,a_i) = &\frac{1}{N_{conf}}(\Sigma_i {L_{conf}(p_i,p^*_i)} \\
  & +\frac{\beta}{N_{loc}}\Sigma_i{p^*_iL_{loc}(t_i,g_i,a_i)}),
  \end{aligned}
  \label{eq:e2}
\end{equation}
where $N_{conf}$ and $N_{loc}$ indicate the number of positive and negative anchors, and the number of positive anchors respectively, $L_{conf}$ is the softmax loss over two classes (face vs. background), and $L_{loc}$ is the smooth $L_1$ loss between the parameterizations of the predicted box $t_i$ and ground-truth box $g_i$ using the anchor $a_i$.
When $p_i^*=1$ ($p_i^*=\{0,1\}$), the anchor $a_i$ is positive and the localization loss is activated.
$\beta$ is a weight to balance the effects of the two terms.
Compared to the enhanced feature maps in the same level, the original feature maps have less semantic information for classification but more high resolution location information for detection.
Therefore, we believe that the original feature maps can detect and classify smaller faces.
As the result, we propose the First Shot multi-task Loss with a set of smaller anchors as follows:
\begin{equation}\
%\small
\begin{aligned}
  \mathcal{L}_{FSL}(p_i,p^*_i,t_i,g_i,sa_i) = &\frac{1}{N_{conf}}\Sigma_i {L_{conf}(p_i,p^*_i)} \\
  & + \frac{\beta}{N_{loc}}\Sigma_i{p^*_iL_{loc}(t_i,g_i,sa_i)},
  \end{aligned}
  \label{eq:e3}
\end{equation}
where $sa$ indicates the smaller  anchors in the first  shot layers, and the two shots losses can be weighted summed into a whole Progressive Anchor Loss as follows:
\begin{equation}\
%\small
\begin{aligned}
  \mathcal{L}_{PAL} = \mathcal{L}_{FSL}(sa)+\lambda\mathcal{L}_{SSL}(a).
  \end{aligned}
  \label{eq:e4}
\end{equation}
Note that anchor size in the first shot is half of ones in the second shot, and $\lambda$ is weight factor. Detailed assignment on the anchor size is described in Sec.~\ref{section:3.4}.
In prediction process, we only use the output of the second shot, which means no additional computational cost is introduced.

\begin{table}[t!]
%\small
\footnotesize
%\scriptsize
\centering
%  \resizebox{0.98\linewidth}{!}{
%\begin{center}
\caption{\small The stride size, feature map size, anchor scale, ratio, and number of six original/enhanced features for two shots.}
\label{table:tab1}\figvspace
\begin{tabular}{c|c|c|c|c|c}
%\begin{tabular}{|p{1cm}|p{0.5cm}|p{1.5cm}|p{1.5cm}|p{1.5cm}|p{1.5cm}|p{1.5cm}|p{1.5cm}|p{1.5cm}||p{1.6cm}|}
    \hline
    Feature  & Stride & Size  & Scale & Ratio & Number \\
    \hline
    ef$\_$1 (of$\_$1) & $4$ &  $160\times 160$ &     $16$ ($8$)  &     $1.5:1$  &   $25600$ \\
    ef$\_$2 (of$\_$2) &   $8$    &    $80\times80$      &   $32$ ($16$)  &   $1.5:1$   &   $6400$ \\
    ef$\_$3 (of$\_$3) &   $16$    &   $40\times40$      &   $64$ ($32$)   &  $1.5:1$   &  $1600$ \\
    ef$\_$4 (of$\_$4) &   $32$    &   $20\times20$     &   $128$ ($64$)  &  $1.5:1$   &   $400$ \\
    ef$\_$5 (of$\_$5) &   $64$    &   $10\times10$     &   $256$ ($128$) & $1.5:1$   &  $100$ \\
    ef$\_$6 (of$\_$6) &  $128$    &   $5\times5$       &   $512$ ($256$) &  $1.5:1$   &   $25$ \\
    \hline
\end{tabular}
%}
 \vspace{-2mm}
\end{table}

\subsection{Improved Anchor Matching}\label{section:3.4}
%During training, we need to compute positive and negative anchors and determine which anchor corresponds to its face bounding box.
Current anchor matching method is bidirectional between the anchor and ground-truth face.
Therefore, anchor design and face sampling during augmentation are collaborative to match the anchors and faces as far as possible for better initialization of the regressor.
Our IAM targets on addressing the contradiction between the discrete anchor scales and continuous face scales, in which the faces are augmented by $S_{input}*S_{face}/S_{anchor}$ ($S$ indicates the spatial size) with the probability of $40\%$ so as to \textit{increase} the positive anchors, \textit{stabilize} the training and thus improve the results.
Table~\ref{table:tab1} shows details of our anchor design on how each feature map cell is associated to the fixed shape anchor.
We set anchor ratio $1.5$:$1$ based on face scale statistics.
Anchor size for the original feature is one half of the enhanced feature. Additionally, with probability of $2/5$, we utilize anchor-based sampling like data-anchor-sampling in PyramidBox, which randomly selects a face in an image, crops sub-image containing the face, and sets the size ratio between sub-image and selected face to $640$/rand ($16,32,64,128,256,512$). For the remaining $3/5$ probability, we adopt data augmentation similar to SSD~\cite{liu2016ssd}.
In order to improve the recall rate of faces and ensure anchor classification ability simultaneously, we set Intersection-over-Union (IoU) threshold $0.4$ to assign anchor to its ground-truth faces.

%------------------------------------------------------------------------
\section{Experiments}\label{section:4}

\subsection{Implementation Details}\label{section:4.0}
First, we present the details in implementing our network.
The backbone networks are initialized by the pretrained VGG/ResNet on ImageNet.
All newly added convolution layers' parameters are initialized by the `xavier' method.
We use SGD with $0.9$ momentum, $0.0005$ weight decay to fine-tune our DSFD model.
The batch size is set to $16$.
The learning rate is set to $10^{-3}$ for the first $40$k steps, and we decay it to $10^{-4}$ and $10^{-5}$ for two $10$k steps.

During inference, the first shot's outputs are ignored and the second shot predicts top $5$k high confident detections.
Non-maximum suppression is applied with jaccard overlap of $0.3$ to produce top $750$ high confident bounding boxes per image.
For $4$ bounding box coordinates, we round down top left coordinates and round up width and height to expand the detection bounding box.
The official code has been released at: \url{https://github.com/TencentYoutuResearch/FaceDetection-DSFD}.

\begin{table}[t!]
\small
%\footnotesize
%\scriptsize
\centering
%  \resizebox{0.98\linewidth}{!}{
%\begin{center}
\caption{\small Effectiveness of Feature Enhance Module on the AP performance.}
\label{table:tab2}\figvspace
\begin{tabular}{c|c|c|c}
%\begin{tabular}{|p{1cm}|p{0.5cm}|p{1.5cm}|p{1.5cm}|p{1.5cm}|p{1.5cm}|p{1.5cm}|p{1.5cm}|p{1.5cm}||p{1.6cm}|}
    \hline
    Component  & Easy & Medium  & Hard\\
    \hline
    FSSD+VGG16           &        $92.6\%$     &    $90.2\%$    &   $79.1\%$\\
    FSSD+VGG16+FEM       &        $\textbf{93.0}\%$        &   $\textbf{91.4}\%$    &   $\textbf{84.6}\%$ \\
    \hline
\end{tabular}
%}
\end{table}

\begin{table}[t!]
\small
%\footnotesize
%\scriptsize
\centering
%  \resizebox{0.98\linewidth}{!}{
%\begin{center}
\caption{\small Effectiveness of Progressive Anchor Loss on the AP performance.}
\label{table:tab3}\figvspace
\begin{tabular}{c|c|c|c}
%\begin{tabular}{|p{1cm}|p{0.5cm}|p{1.5cm}|p{1.5cm}|p{1.5cm}|p{1.5cm}|p{1.5cm}|p{1.5cm}|p{1.5cm}||p{1.6cm}|}
    \hline
    Component  & Easy & Medium  & Hard\\
    \hline
    FSSD+RES50              &           $93.7\%$     &    $92.2\%$   &   $81.8\%$  \\
    FSSD+RES50+FEM         &      $95.0\%$     &    $94.1\%$   &   $88.0\%$  \\
    FSSD+RES50+FEM+PAL    & $\textbf{95.3}\%$    &     $\textbf{94.4}\%$   &   $\textbf{88.6}\%$  \\
    \hline
\end{tabular}
%}
\end{table}

\begin{figure}[t]
  \centering
  \includegraphics[trim={0 0 0 0mm},clip,width=1\linewidth]{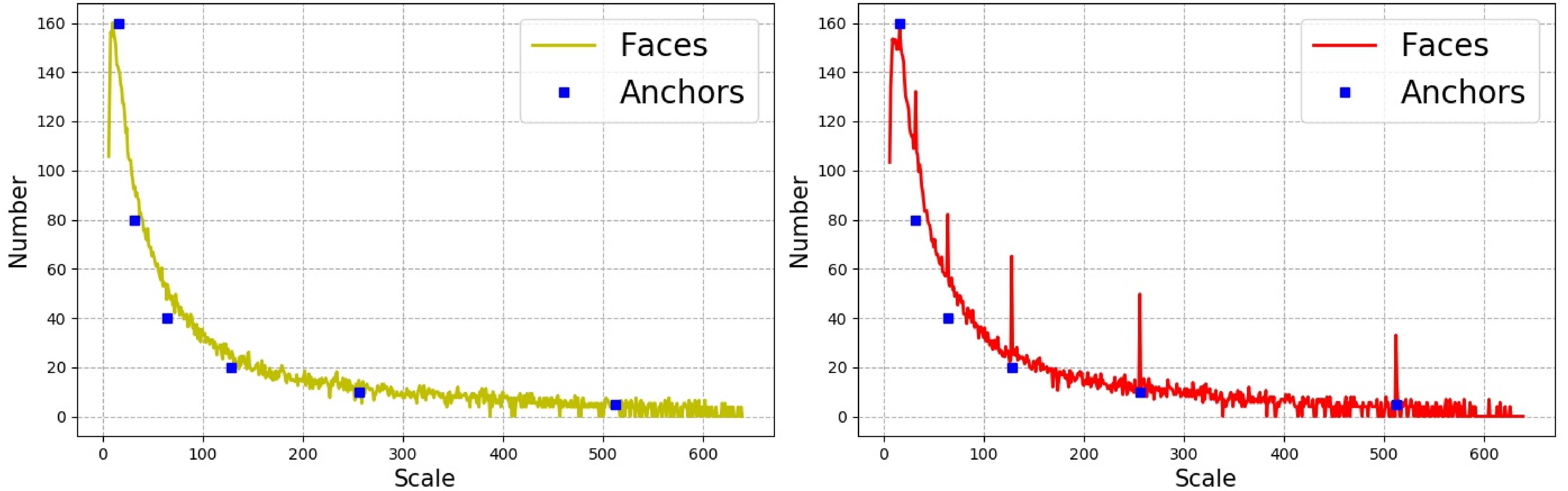}
  \vspace{-7mm}
  \caption{\small \textbf{The number distribution of different scales of faces} compared between traditional anchor matching (\textbf{Left}) and our improved anchor matching (\textbf{Right}).
  } %FIXME
  \label{fig:comp_anchor1} \figvspace
\end{figure}

\vspace{-1mm}
\subsection{Analysis on DSFD}\label{section:4.1}
In this subsection, we conduct extensive experiments and ablation studies on the WIDER FACE dataset to evaluate the effectiveness of several contributions of our proposed framework, including feature enhance module, progressive anchor loss, and improved anchor matching.
%We conduct a set of ablation experiments on the WIDER FACE dataset to analyze our model in detail.
For fair comparisons, we use the same parameter settings for all the experiments, except for the specified changes to the components.
All models are trained on the WIDER FACE training set and evaluated on validation set.
To better understand DSFD, we select different baselines to ablate each component on how this part affects the final performance.

\Paragraph{Feature Enhance Module}
First, We adopt anchor designed in S3FD~\cite{zhang2017s}, PyramidBox~\cite{tang2018pyramidbox} and six original feature maps generated by VGG$16$ to perform classification and regression, which is named Face SSD (FSSD) as the baseline.
We then use VGG$16$-based FSSD as the baseline to add feature enchance module for comparison.
Table~\ref{table:tab2} shows that our feature enhance module can improve VGG16-based FSSD from $92.6\%$, $90.2\%$, $79.1\%$ to $93.0\%$, $91.4\%$, $84.6\%$.

\Paragraph{Progressive Anchor Loss}
Second, we use Res$50$-based FSSD as the baseline to add progressive anchor loss for comparison. We use four residual blocks' ouputs in ResNet to replace the outputs of conv$3\_3$, conv$4\_3$, conv$5\_3$, conv$\_$fc$7$ in VGG.
Except for VGG16, we do not perform ¡®layer normalization¡¯.
Table~\ref{table:tab3} shows our progressive anchor loss can improve Res50-based FSSD using FEM from $95.0\%$, $94.1\%$, $88.0\%$ to $95.3\%$, $94.4\%$, $88.6\%$.

\begin{figure}[t]
  \centering
  \includegraphics[trim={0 0 0 0mm},clip,width=0.75\linewidth]{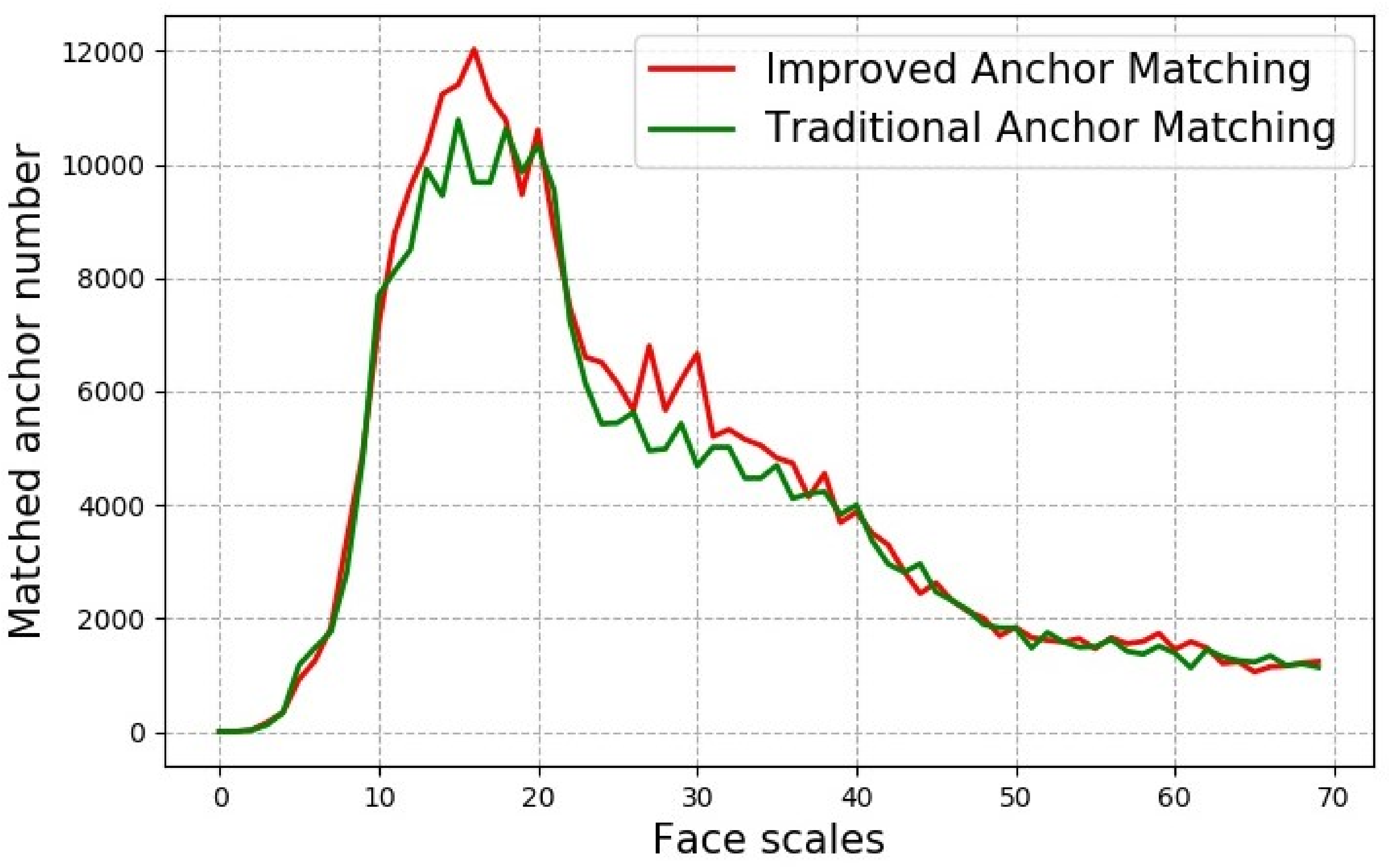}
  \vspace{-4mm}
  \caption{\small \textbf{Comparisons on number distribution of matched anchor for ground truth faces} between traditional anchor matching (blue line) and our improved anchor matching (red line).
  we actually set the IoU threshold to $0.35$ for the traditional version.
  That means even with a higher threshold (\textit{i.e.}, $0.4$), using our IAM, we can still achieve more matched anchors.
  Here, we choose a slightly higher threshold in IAM so that to better balance the \textit{number} and \textit{quality} of the matched faces.
  } %FIXME
  \label{fig:comp_anchor2} \figvspace
\end{figure}

\begin{figure*}[t]
  \centering
  \includegraphics[trim={0 0 0 0mm},clip,width=0.86\linewidth]{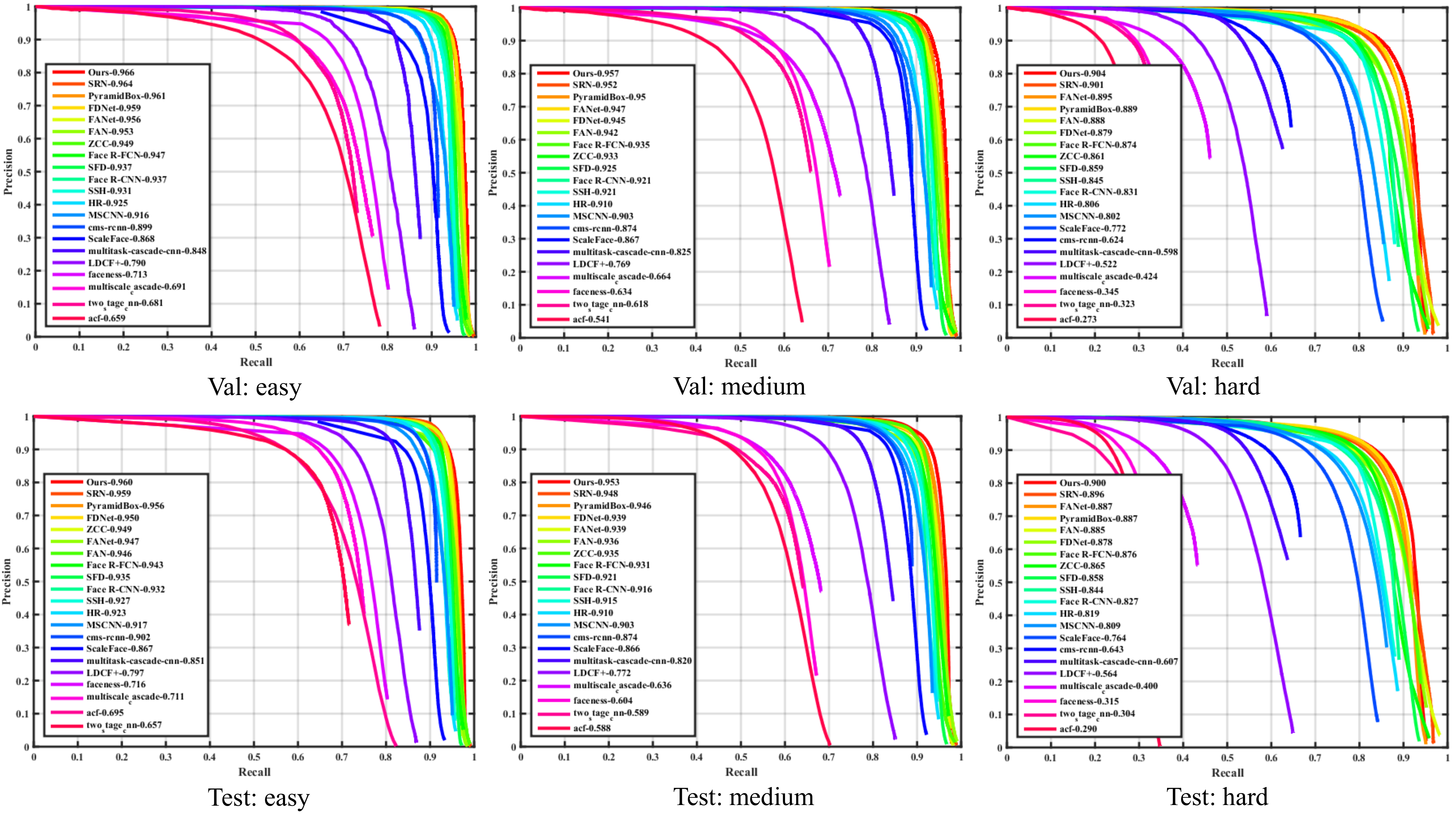}
  \vspace{-4mm}
  \caption{\small \textbf{Precision-recall curves on WIDER FACE} validation and testing subset.
  }
  \label{fig:widerface} \figvspace
\end{figure*}

\begin{table*}[t!]
%\small
%\footnotesize
%\tiny
\centering
%  \resizebox{0.98\linewidth}{!}{
%\begin{center}
 \vspace{2mm}
\caption{ Effectiveness of Improved Anchor Matching on the AP performance.}
\label{table:tab4}\figvspace
\begin{tabular}{c|c|c|c}
%\begin{tabular}{|p{1cm}|p{0.5cm}|p{1.5cm}|p{1.5cm}|p{1.5cm}|p{1.5cm}|p{1.5cm}|p{1.5cm}|p{1.5cm}||p{1.6cm}|}
    \hline
    Component  & Easy & Medium  & Hard \\
    \hline
    FSSD+RES101               &      $95.1\%$  &  $93.6\%$  &  $83.7\%$ \\
    FSSD+RES101+FEM           &      $95.8\%$  &  $95.1\%$  &  $89.7\%$  \\
    FSSD+RES101+FEM+IAM       &      $96.1\%$  &  $95.2\%$  &  $90.0\%$  \\
    FSSD+RES101+FEM+IAM+PAL   &      $96.3\%$  &  $95.4\%$   &  $90.1\%$ \\
    FSSD+RES152+FEM+IAM+PAL   &   $\textbf{96.6}\%$ &  $\textbf{95.7}\%$  & $90.4\%$ \\
    FSSD+RES152+FEM+IAM+PAL+LargeBS  &   $96.4\%$ &  $\textbf{95.7}\%$ &  $\textbf{91.2}\%$ \\
    \hline
\end{tabular}
%}
 \vspace{-2mm}
\end{table*}

\begin{table*}[t!]
%\small
%\footnotesize
%\tiny
\centering
%  \resizebox{0.98\linewidth}{!}{
%\begin{center}
 \vspace{2mm}
\caption{Effectiveness of different backbones.}
\label{table:tab5}\figvspace
\begin{tabular}{c|c|c|c|c|c}
%\begin{tabular}{|p{1cm}|p{0.5cm}|p{1.5cm}|p{1.5cm}|p{1.5cm}|p{1.5cm}|p{1.5cm}|p{1.5cm}|p{1.5cm}||p{1.6cm}|}
    \hline
    Component  & Params & ACC@Top-$1$  & Easy & Medium & Hard\\
    \hline
    FSSD+RES101+FEM+IAM+PAL  &  $399$M & $77.44\%$ & $96.3\%$  &  $95.4\%$ & $90.1\%$ \\
    FSSD+RES152+FEM+IAM+PAL  &  $459$M & $78.42\%$ & $\bf 96.6\%$  &  $\bf 95.7\%$ & $\bf 90.4\%$ \\
    FSSD+SE-RES101+FEM+IAM+PAL  &  $418$M & $78.39\%$ & $95.7\%$  &  $94.7\%$ & $88.6\%$  \\
    FSSD+DPN98+FEM+IAM+PAL  &  $515$M & $79.22\%$ & $96.3\%$  &  $95.5\%$ & $\bf 90.4\%$  \\
    FSSD+SE-RESNeXt101$\_32$$\times$$4$d+FEML+IAM+PA  &  $416$M & $80.19\%$ & $95.7\%$  &  $94.8\%$ & $88.9\%$ \\
    \hline
\end{tabular}
%}
 \vspace{-2mm}
\end{table*}

\begin{figure*}[t]
  \centering
  \includegraphics[trim={0 0 0 0mm},clip,width=0.89\linewidth]{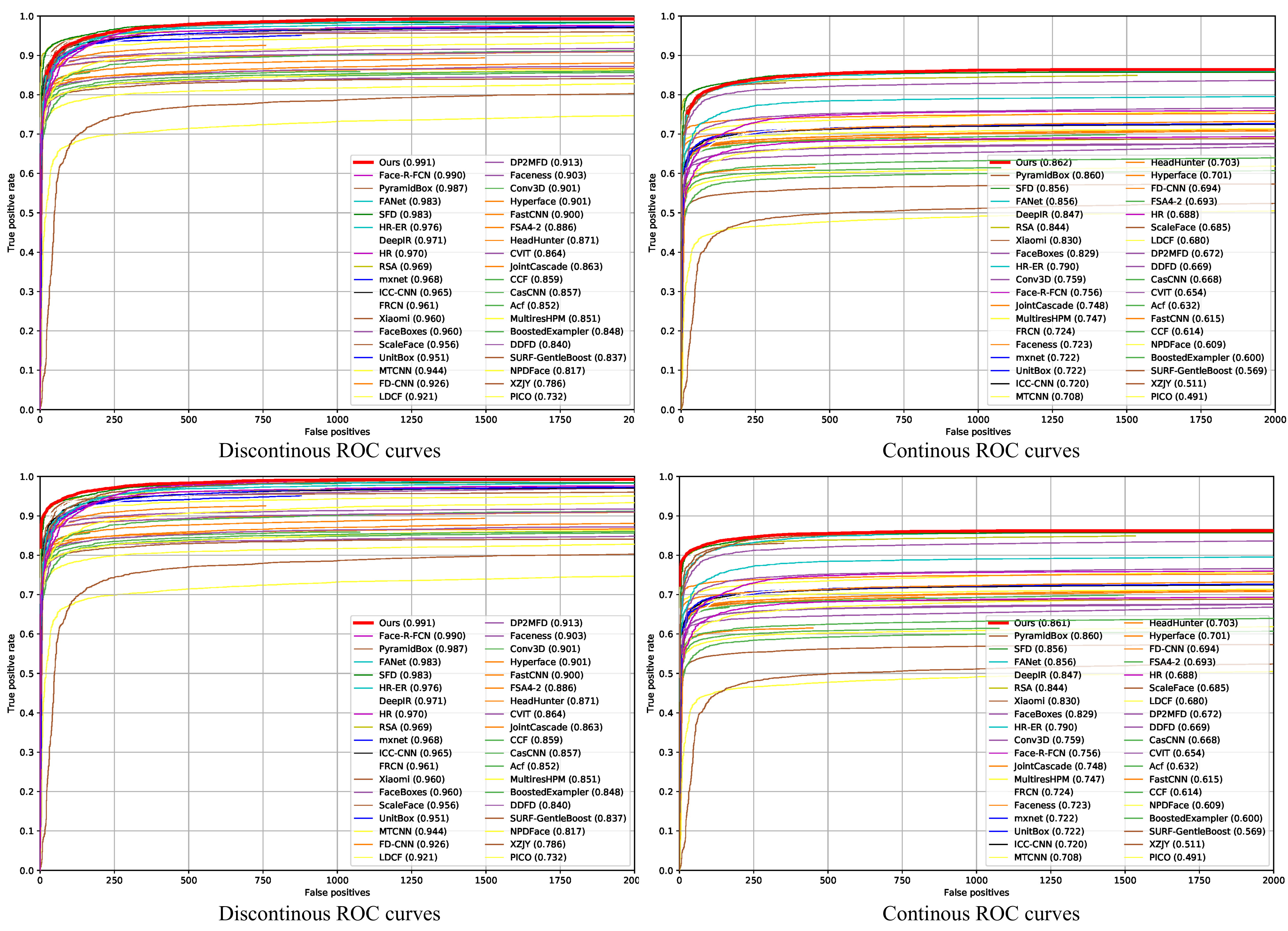}
  \vspace{-4mm}
  \caption{\small \textbf{Comparisons with popular state-of-the-art methods on the FDDB dataset}.
  The first row shows the ROC results without additional annotations, and the second row shows the ROC results with additional annotations.
  }
  \label{fig:fddb} \figvspace
\end{figure*}

\begin{table}[t]
\begin{center}
\caption{\small \textbf{FEM vs.~RFB} on WIDER FACE.}
\label{table:comp_RFB}\vspace{-4mm}
\footnotesize
\resizebox{0.98\linewidth}{!}{
\begin{tabular}{l|c|c|c}
\hline
Backbone - ResNet$101$ ($\%$) & Easy  & Medium & Hard  \\ \hline
DSFD (RFB)  & $96.0$ & $94.5$ & $87.2$ \\
DSFD (FPN) / (FPN+RFB)  & $96.2$ / $96.2$ & $95.1$ / $95.3$ & $89.7$ / $89.9$ \\
%DSFD ()  & $96.2$ & $95.3$ & $89.9$ \\
DSFD (FEM)  & $\textbf{96.3}$ & $\textbf{95.4}$ & $\textbf{90.1}$ \\ \hline
\end{tabular}}\figvspace\figvspace
\end{center}
\end{table}

\Paragraph{Improved Anchor Matching}
To evaluate our improved anchor matching strategy, we use Res$101$-based FSSD without anchor compensation as the baseline.
Table~\ref{table:tab4} shows that our improved anchor matching can improve Res101-based FSSD using FEM from $95.8\%$, $95.1\%$, $89.7\%$ to $96.1\%$, $95.2\%$, $90.0\%$.
Finally, we can improve our DSFD to $96.6\%$, $95.7\%$, $90.4\%$ with ResNet$152$ as the backbone.

Besides, Fig.~\ref{fig:comp_anchor1} shows that our improved anchor matching strategy greatly increases the number of ground truth faces that are closed to the anchor, which can reduce the contradiction between the discrete anchor scales and continuous face scales.
Moreover, Fig.~\ref{fig:comp_anchor2} shows the number distribution of matched anchor number for ground truth faces, which indicates our improved anchor matching can significantly increase the matched anchor number, and the averaged number of matched anchor for different scales of faces can be improved from $6.4$ to about $6.9$.

\Paragraph{Comparison with RFB}
Our FEM differs from RFB in two aspects.
First, our FEM is based on FPN to make full use of feature information from different spatial levels, while RFB ignores.
Second, our FEM adopts stacked dilation convolutions in a multi-branch structure, which efficiently leads to larger Receptive Fields (RF) than
RFB that only uses one dilation layer in each branch, \textit{e.g.}, $R^3$ in FEM compared to $R$ in RFB where indicates the RF of one dilation convolution.
Tab.~\ref{table:comp_RFB} clearly demonstrates the superiority of our FEM over RFB, even when RFB is equipped with FPN.

From the above analysis and results, some promising conclusions can be drawn:
$1$) Feature enhance is crucial. We use a more robust and discriminative feature enhance module to improve the feature presentation ability, especially for hard face.
$2$) Auxiliary loss based on progressive anchor is used to train all $12$ different scale detection feature maps, and it improves the performance on easy, medium and hard faces simultaneously.
$3$) Our improved anchor matching provides better initial anchors and ground-truth faces to regress anchor from faces, which achieves the improvements of $0.3\%$, $0.1\%$, $0.3\%$ on three settings, respectively.
Additionally, when we enlarge the training batch size (\emph{i.e.}, LargeBS), the result in hard setting can get 91.2$\%$ AP.

\Paragraph{Effects of Different Backbones}
To better understand our DSFD, we further conducted experiments to examine how different backbones affect classification and detection performance.
Specifically, we use the same setting except for the feature extraction network, we implement SE-ResNet$101$, DPN$-98$, SE-ResNeXt101$\_32$$\times$$4$d following the ResNet$101$ setting in our DSFD.
From Table~\ref{table:tab5}, DSFD with SE-ResNeXt101$\_32$$\times$$4$d got $95.7\%$, $94.8\%$, $88.9\%$, on easy, medium and hard settings respectively, which indicates that more complexity model and higher Top-$1$ ImageNet classification accuracy may not benefit face detection AP.
Therefore, in our DSFD framework, better performance on classification are not necessary for better performance on detection, which is consistent to the conclusion claimed in~\cite{li2018detnet,Huang2017speed}.
Our DSFD enjoys high inference speed benefited from simply using the second shot detection results.
For VGA resolution inputs to Res$50$-based DSFD, it runs $22$ FPS on NVIDA GPU P$40$ during inference.

%Moreover, we adopt several methods to further improve the detection performance, including expanding the detection coordinates (\emph{i.e.}, Expansion) and adding more convolutional layers for classification and regression branches simultaneously (\emph{i.e.}, DeepHead), which improves our DSFD to $96.6\%$, $95.7\%$, $90.4\%$. Additionally, when we enlarge the training batch size (\emph{i.e.}, LargeBS), the result in hard setting can get $91.0\%$ AP.

\begin{figure*}[t]
  \centering
  \includegraphics[trim={0 0 0 0mm},clip,width=0.84\linewidth]{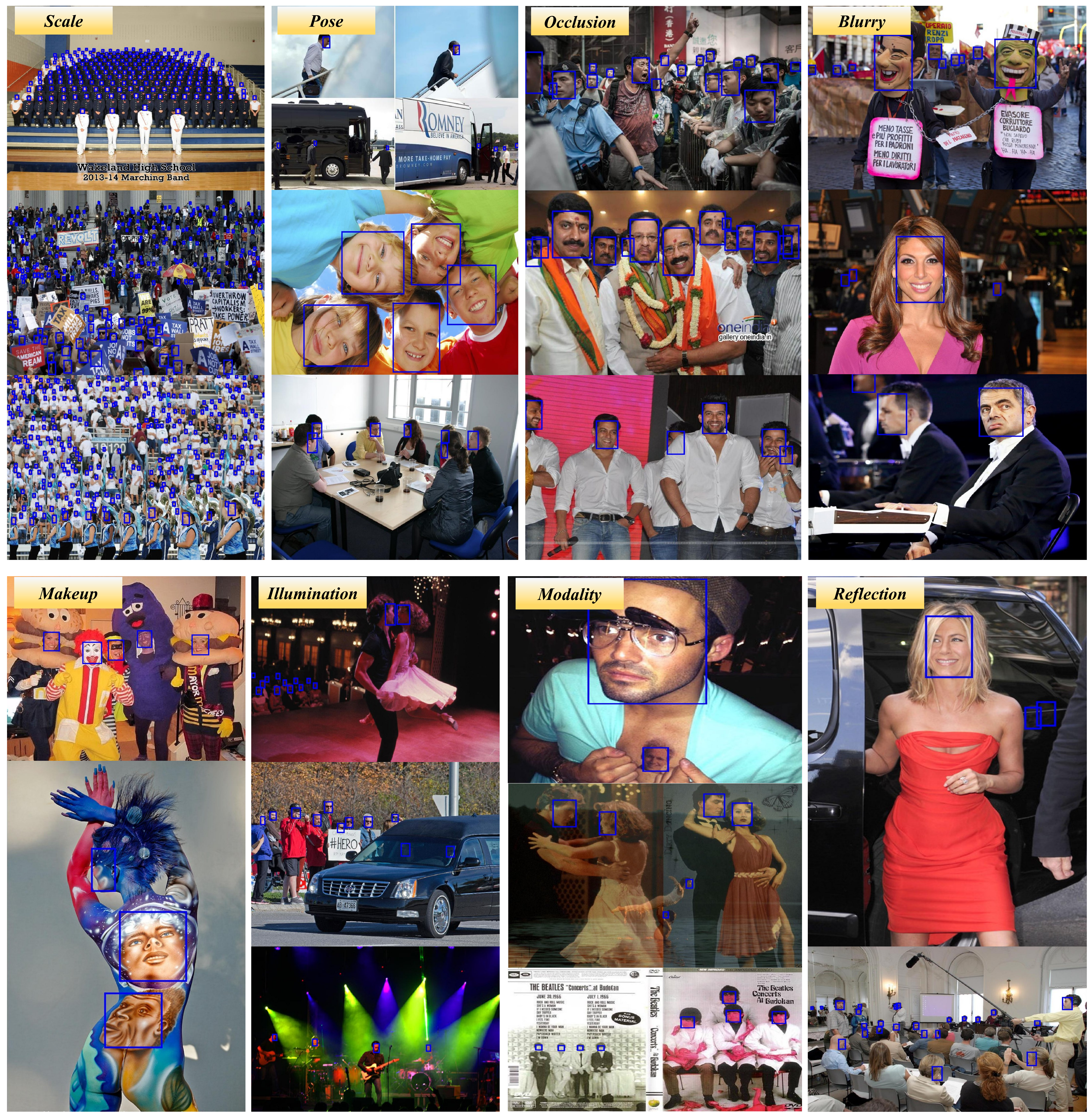}
  \vspace{-3mm}
  \caption{\small \textbf{Illustration of our DSFD to various large variations} on scale, pose, occlusion, blurry, makeup, illumination, modality and reflection.
  Blue bounding boxes indicate the detector confidence is above $0.8$.
  } %FIXME
  \label{fig:largevariations} \figvspace
\end{figure*}

\subsection{Comparisons with State-of-the-Art Methods}\label{section:4.2}
We evaluate the proposed DSFD on two popular face detection benchmarks, including WIDER FACE~\cite{yang2016wider} and Face Detection Data Set and Benchmark (FDDB)~\cite{jain2010fddb}.
Our model is trained only using the training set of WIDER FACE, and then evaluated on both benchmarks without any further fine-tuning.
We also follow the similar way used in~\cite{wang2017face} to build the image pyramid for multi-scale testing and use more powerful backbone similar as~\cite{chi2018selective}.

\Paragraph{WIDER FACE Dataset}
It contains $393,703$ annotated faces with large variations in scale, pose and occlusion in total $32,203$ images. For each of the $60$ event classes, $40\%$, $10\%$, $50\%$ images of the database are randomly selected as training, validation and testing sets. Besides, each subset is further defined into three levels of difficulty: 'Easy', 'Medium', 'Hard' based on the detection rate of a baseline detector.
As shown in Fig.~\ref{fig:widerface}, our DSFD achieves the best performance among all of the state-of-the-art face detectors based on the average precision (AP) across the three subsets, \emph{i.e.}, $96.6\%$ (Easy), $95.7\%$ (Medium) and $90.4\%$ (Hard) on validation set, and $96.0\%$ (Easy), $95.3\%$ (Medium) and $90.0\%$ (Hard) on test set.
Fig.~\ref{fig:largevariations} shows more examples to demonstrate the effects of DSFD on handling faces with various variations, in which the blue bounding boxes indicate the detector confidence is above $0.8$.

\vspace{-1mm}
\Paragraph{FDDB Dataset}
It contains $5,171$ faces in $2,845$ images taken from the faces in the wild data set.
Since WIDER FACE has bounding box annotation while faces in FDDB are represented by ellipses, we learn a post-hoc ellipses regressor to transform the final prediction results.
As shown in Fig.~\ref{fig:fddb}, our DSFD achieves state-of-the-art performance on both discontinuous and continuous ROC curves, \emph{i.e.} $99.1\%$ and $86.2\%$ when the number of false positives equals to $1,000$.
After adding additional annotations to those unlabeled faces~\cite{zhang2017s}, the false positives of our model can be further reduced and outperform all other methods.

%
%------------------------------------------------------------------------
\section{Conclusions}
\vspace{-2mm}
This paper introduces a novel face detector named Dual Shot Face Detector (DSFD).
In this work, we propose a novel Feature Enhance Module that utilizes different level information and thus obtains more discriminability and robustness features.
Auxiliary supervisions introduced in early layers by using smaller anchors are adopted to effectively facilitate the features.
Moreover, an improved anchor matching method is introduced to match anchors and ground truth faces as far as possible to provide better initialization for the regressor.
Comprehensive experiments are conducted on popular face detection benchmarks, FDDB and WIDER FACE, to demonstrate the superiority of our proposed DSFD compared with the state-of-the-art face detectors, \textit{e.g.}, SRN and PyramidBox.

%\section{Acknowledgments}
%This work was supported by the National Science Fund of China under Grant Nos. 61876083, U1713208, and Program for Changjiang Scholars.

{\small
\bibliographystyle{ieee_fullname}
\bibliography{egbib}
}

\end{document}